# Detecting Wildfire Flame and Smoke through Edge Computing using Transfer Learning Enhanced Deep Learning Models

Giovanny Vazquez, Shengjie Zhai, and Mei Yang
*Dept. of Electrical & Computer Engineering*
*Univeristy of Nevada, Las Vegas, Las Vegas, Nevada, USA*
Emails: Giovanny.Vazquez@unlv.edu, Shengjie.Zhai@unlv.edu, Mei.Yang@unlv.edu

*Abstract*— **Autonomous unmanned aerial vehicles (UAVs) integrated with edge computing capabilities empower real-time data processing directly on the device, dramatically reducing latency in critical scenarios such as wildfire detection. This study underscores Transfer Learning's (TL) significance in boosting the performance of object detectors for identifying wildfire smoke and flames, especially when trained on limited datasets, and investigates the impact TL has on edge computing metrics. With the latter focusing on how TL-enhanced You Only Look Once (YOLO) models perform in terms of inference time, power usage, and energy consumption when using edge computing devices. This study utilizes the Aerial Fire and Smoke Essential (AFSE) dataset as the target, with the Flame and Smoke Detection Dataset (FASDD) and the Microsoft Common Objects in Context (COCO) dataset serving as source datasets. We explore a two-stage cascaded TL method, utilizing D-Fire or FASDD as initial stage target datasets and AFSE as the subsequent stage. Through fine-tuning, TL significantly enhanced detection precision, achieving up to 79.2% mean Average Precision (mAP@0.5), reduced training time, and increased model generalizability across the AFSE dataset. However, cascaded TL yielded no notable improvements and TL alone did not benefit from the edge computing metrics evaluated. Lastly, this work found that YOLOv5n remains a powerful model when lacking hardware acceleration, finding that YOLOv5n can process images nearly twice as fast as its newer counterpart, YOLO11n. Overall, the results affirm TL's role in augmenting the accuracy of object detectors while also illustrating that additional enhancements are needed to improve edge computing performance.**

## I. INTRODUCTION

With increasing wildfire intensity and frequency comes a heightened environmental, economic, and health impact [1]. Early detection is a crucial tactic in mediating the costs wrought by this form of disaster. Traditionally, this was pursued by physical sensors or visual detection. Physical sensors, however, are impractical on the scale needed to effectively monitor expansive areas [2]. Additionally, past approaches to visual detection using watchtowers or satellites have also proved to have their own drawbacks [3]. Watchtowers suffer from being fixed in place, construction costs while satellites lack the spatial and temporal resolution needed to catch fires as they are first breaking out [3]. As a result, the use of unmanned aerial vehicles (UAVs) serves as a promising alternative to detecting wildfires early [3]. To moderate the need for manual observation, UAV visual detection can be coupled with Deep-Learning (DL) based computer vision objection models.

There are two significant challenges in the deployment of DL-based computer vision algorithms on UAV platforms. The first challenge lies in the requirement for large and diverse datasets of annotated images. The second challenge lies in the implementation of DL-based models on edge computing devices mounted to UAVs for the purpose of real-time processing. Edge computing devices using only Central Processing Units (CPUs) have severely limited computational resources which can drastically reduce detection frame rates. This inability to use a high frame rate can cause the object detection model to miss the processing of frames with potential fire instances. Whilst flying at a slower speed would mitigate this issue, doing so would restrict the area over which a UAV is able to monitor on a given charge due to their limited battery capacity. Furthermore, although edge computing resources can be augmented through cloud computing resources or hardware accelerators – such as graphical processing units (GPUs) or field programmable gate arrays (FPGAs) - these methods carry their own drawbacks.

Cloud computing will introduce communication delays, due to sending complete video for processing, and in addition, can induce network congestion [4][5]. GPUs, although significantly improving computing power, will increase the power consumption of edge devices [6]. FPGAs, although having improved power consumption in addition to improved computing power, lack the pre-built libraries and packages available to GPUs and CPUs making development difficult [6]. Moreover, both GPUs and FPGAs can introduce additional costs for edge devices due to specialized hardware that hamper affordability and in turn limit usability for low-resourced first responders. Accordingly, wildfire object detection models should be optimized for edge computing devices that lack hardware acceleration so that accuracy is maximized while inference time, power usage, and costs are minimized.

The focus of our research is the development of an optimized lightweight convolutional neural network (CNN) model, designed for CPU enabled UAV-based, real-time wildfire detection. Our past work has illustrated the importance of Transfer Learning (TL) in elevating the performance of YOLOv5 [7] for the purpose of aerial based wildfire flame and smoke detection, particularly when limited by small datasets lacking diversity [8]. This work further enhances this exploration through a refined implementation of TL, an expanded comparison to other state-of-the-art object detectors, the investigation of a cascaded TL approach, and an examination of TL's effect on edge computing metrics.

The remainder of this paper is constructed as follows: Section II introduces the datasets and provides an overview of the current state-of-the-art for object detection and its application in detecting wildfire flame and smoke; Section III details the datasets used, model selection, evaluation metrics, experimental setup, and explicates the TL process; Section IV elaborates on the experiments performed, model setup, and

This work is supported in part by the Nevada Space Grant Consortium Research Opportunity Fellowship, NSF under grant no. 1949585, and the UNLV AI SUSTEIN Seed Grant.

findings; and Section V summarizes our work and contributions. Through this research, we emphasize TL's value in augmenting the performance of object detection models limited by small datasets and explore its impact in enhancing edge computing metrics for UAV-based wildfire detection.

## II. RELATED WORK

### A. Relevant Datasets

The Flame and Smoke Detection Dataset (FASDD) emerges as the most exhaustive dataset in the general fire detection domain [9]. Although FASDD is comprised of 95,314 images suitable for standard flame and smoke object detection training, these images contain flame and smoke instances in a wide varying list of scenarios which are often quite different than the instances found in wildfires. For instance, the smoke category includes images where the smoke is generated from a cigarette, burning building, or factory smokestack. Similarly, the fire category includes images where the fire is generated from stoves, candles, or torches. In turn, FASDD contains many non-application specific instances of fire and smoke. In a similar manner, the D-Fire dataset also provides a broad collection of fire and smoke images [10]. Although, it is worth highlighting that D-Fire is much smaller than FASDD and from visual inspection, provides images more targeted to large urban fires or wildfires. The most relevant collection of wildfire flame and smoke instances are provided by the Fire Luminosity Airborne-based Machine Learning Evaluation (FLAME) datasets, 1 and 2 [11][12]. However, both FLAME datasets contain many similar frames resulting in datasets that lack diversity, primarily due to their compilation from singular recordings of prescribed burns in Northern Arizona [11][12]. Consequently, the lack of datasets tailored to aerial-based wildfire detection is what motivated this work to generate a new dataset which is detailed further in Section III.

### B. Object Detection State-of-the-Art

The realm of object detection algorithms is broadly categorized into either CNN-based models or transformer-based models [13]. CNN-based models are further divided into either one-stage or two-stage models. Two-stage models differ from one-stage in that they first generate candidate bounding boxes from region of interest (RoI) proposals and then extract features from these bounding boxes. One-stage models, on the other hand, implement both these steps in a single stage [13]. While one-stage algorithms are lauded for their speed, two-stage algorithms are celebrated for their precision [14]. Transformer-based models are also further categorized into either end-to-end or Vision Transformer (ViT) models [13]. End-to-end, also called Detection Transformer (DETR), models utilize the encoder-decoder module of transformers and are often used in detection. ViT models divide images into patches that are then stacked into a vector to be used as inputs to the model and are often used in classification. Note that DETR models remove the need for hand-crafted processes, such as anchor boxes and non-maximum suppression (NMS), simplifying the architecture of object detectors [13].

Both CNN and transformer-based detectors have seen an influx of methods over the last decade. State-of-the-art (SOTA) variants for multi-stage detectors include Cascade region-based convolutional neural networks (Cascade R-CNN) [15] and Dynamic-RCNN [16]. One way previous two-stage detectors, such as Faster R-CNN [17], were limited was in their generation of proposals. As detailed in [15], training with fixed high accuracy thresholds, although producing less noisy bounding boxes, yields degraded detection performance due to a reduced number of proposals to train on. To address this, Cascade-RCNN refined proposals in several stages, ultimately leaving many proposals to train on initially while still being able to train on high accuracy examples eventually. Nevertheless, as described in [16], the solution implemented by Cascade-RCNN was time-consuming. To mitigate this, Dynamic-RCNN took advantage of the fact that proposals generated during training would improve without additional modification [16]. By doing so, Dynamic-RCNN was able to implement a similar strategy to Cascade-RCNN without the same added overhead [16].

Among the most prominent one-stage detectors, and the focus of this paper, are the You Only Look Once (YOLO) variants, renowned for their optimal balance of speed and accuracy [2][3]. However, other one-stage methods have shown promise with notable methods including Real-Time Models for object Detection (RTM-DET) [18] and Task-aligned One-stage Object Detection (TOOD) [19]. RTM-DET's improvements were two-fold. First, RTM-DET implemented large-kernel depth-wise convolutions to improve contextual modeling [18]. Second, RTM-DET utilized soft labels when matching ground truth boxes to model predictions to improve model accuracy [18]. As for TOOD, this approach enhanced traditional one-stage detectors by addressing the misalignment in prediction that occurs between classification and localization heads [19] TOOD's solution to this problem was to a develop a novel detection head, to learn alignment between classification and localization tasks, while also attempting to pull the anchors for each task closer together [19].

As for SOTA DETR-based models, variants include DETR with improved denoising anchor boxes (DINO) [20] and Dynamic Anchor Box DETR (DAB-DETR) [21]. Initially, DETR models let each query serve as a positional query which inadvertently resulted in several concentration centers that made it difficult to find multiple objects in an image [21]. Conditional DETR resolved this by using explicit positional queries, but this approach was unable to take scale into account [21]. DAB-DETR addressed this limitation by using 4D box coordinates as queries that account for position and scale [21]. DINO continued to improve the DETR architecture by building on the work of DAB-DETR and Denoising-DETR (DN-DETR) [22]. Where DN-DETR was implemented to speed up training convergence of DETR models by using a denoising training method [20].

Overall, DINO was able to make three key contributions that built upon these prior models. First, it improved upon DN-DETR by using a novel *contrastive denoising training* technique that limited duplicate outputs [20]. Second, it improved upon DAB-DETR by using a novel *mixed query selection* method to better initialize queries. Lastly, a novel technique titled *look forward twice* was developed to improve the box prediction of earlier layers by using box prediction information of later layers [20]. A summary of the major contributions for each method explored in this paper is provided in TABLE I.

TABLE I. MAJOR CONTRIBUTIONS FOR VARIOUS SOTA OBJECT DETECTION MODELS

| Category | Method | Contributions |
|---|---|---|
| One-Stage Detectors | YOLOv5 | Utilizes PyTorch framework and a Focus structure with CSPdarknet53 as the backbone [23] |
| | YOLOv6 | Utilizes a new backbone, a decoupled head, and new classification/regression loss functions [24] |
| | YOLOv8 | Utilizes anchor-free detectors [24] |
| | YOLOv9 | Utilizes novel lightweight network architecture and programmable gradient information [25] |
| | YOLOv10 | Removes non-maximum suppression and adds large-kernel convolution / partial self-attention modules [26] |
| | YOLO11 | Utilizes Cross-Stage Partial with Self-Attention module and replaces C2f block with C3k2 block for efficiency/accuracy improvements [27] |
| | TOOD | Utilizes novel head structure and alignment-oriented learning approach to enhance interaction between classification and localization tasks [19] |
| | RTM-DET | Utilizes large-kernel depth-wise convolutions and dynamic soft label assignment [18] |
| Two-Stage Detectors | Dynamic-RCNN | Utilizes automatic adjustment of accuracy threshold and regression loss function [16] |
| | Cascade-RCNN | Utilizes sequence of detectors with increasing accuracy thresholds [15] |
| Transformer-Based Detectors | DAB-DETR | Utilizes dynamically updated anchor boxes as queries in Transformer decoder [21] |
| | DINO | Utilizes novel techniques for denoising training, query initialization, and box prediction [20] |

## III. MATERIALS AND METHODS

### A. Datasets Utilized

The Aerial Fire and Smoke Essential (AFSE) dataset, first developed for our previous work in [8], is again utilized as the primary target dataset. A sample of the images within the AFSE dataset are shown in Fig. 1. AFSE serves to provide a small collection of images comprised of wildfire flame and smoke instances from an aerial perspective. In total, AFSE incorporates 282 images, with no augmentations, containing scenarios with only smoke, fire and smoke, and no fire nor smoke. Additional datasets utilized in this work are the FASDD dataset, the Microsoft Common Objects in Context (COCO) dataset [28], and the D-Fire dataset.

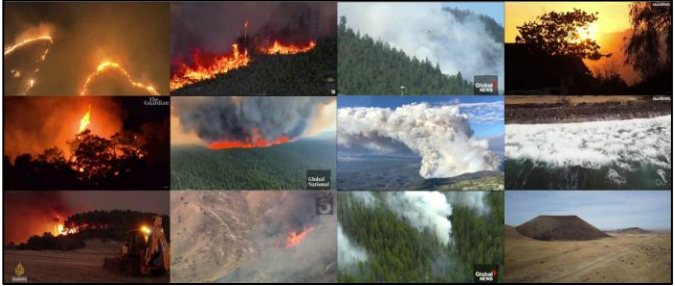

Fig. 1. Image samples from AFSE dataset [8]

### B. Experiment Setup

TABLE II. details the edge computing device used while TABLE III. describes the configuration of the GPU server.

TABLE II. CONFIGURATION OF EDGE COMPUTING DEVICE

| Edge Computing Device Configuration | Details |
|---|---|
| Device | Raspberry Pi 5 – 8GB RAM |
| Deployment environment | Python 3.11 |
| Operating system | Raspberry Pi OS |
| Central processing unit (CPU) | Arm Cortex-A76 64-bit quad-core @ 2.4GHz |

TABLE III. CONFIGURATION OF GPU SERVER

| Server Configuration | Details |
|---|---|
| Deployment environment | Python 3.8 |
| Operating system | Ubuntu 18.04.4 |
| Deep learning framework | PyTorch 1.4.0 |
| Accelerated computing architecture | CUDA 10.0.130 |
| Graphic processing unit (GPU) | Quadro RTX 6000, 24GB VRAM * 8 |
| Central processing unit (CPU) | Intel(R) Xeon(R) Gold 5218 @ 2.30GHz |

### C. Evaluation Metrics

Within this paper, *Average Precision* (AP) and *mean Average Precision* (mAP) serve as the primary metrics to evaluate model accuracy. These metrics are built upon the use of *Precision* and *Recall*. The calculation of Precision and Recall require the use of the following four parameters:

- True Positive (TP): Total number of predictions correctly identified as positive instances.
- True Negative (TN): Total number of predictions correctly identified as negative instances.
- False Positive (FP): Total number of predictions incorrectly identified as positive instances.
- False Negative (FN): Total number of predictions incorrectly identified as negative instances.

Recall measures how well a model can find all the positive class instances. In turn, a model with high Recall produces few false negatives whereas a model with low Recall produces many false negatives. Precision can then be used to quantify how accurate positive class predictions are. A model with low precision will produce many false positives whereas a model with high precision with have few false positives. Correspondingly, Precision and Recall are given by Eqn. (1) and Eqn. (2), respectively.

$$Precision = \frac{TP}{TP+FP} \quad (1)$$

$$Recall = \frac{TP}{TP+FN} \quad (2)$$

Predicted anchor boxes and classes are compared against ground-truth bounding boxes and classes to determine whether a prediction is correct. Classes are directly compared whereas predicted anchor boxes and ground-truth boxes are compared using an Intersection over Union (IoU) threshold. IoU is calculated by taking the ratio of the intersection area between the predicted box and ground-truth box over the union area of the same two boxes. As a result, a correct prediction is one in which the placement of the predicted box over the ground-truth box yields an IoU that surpasses a certain threshold, and one in which the class is correctly identified. For this work, an IoU of 0.5 was utilized. Using varying confidence thresholds, Precision can be plotted against Recall to form a Precision-Recall curve, $p(r)$. The AP is found by taking the area under this curve, as shown in Eqn. (3). The mAP of a model is then found by taking the AP for each class and averaging the results, as summarized in Eqn. (4).

$$Average\ Precision\ (AP) = \int_{r=0}^{1} p(r)dr \quad (3)$$

$$mean\ Average\ Precision\ (mAP) = \frac{1}{k}\sum_{i}^{k} AP_i \quad (4)$$

In addition to AP and mAP, model performance is evaluated on inference speed, average power used during inference, and a normalized energy-delay product (EDP). Inference speed captures the number of images an object detection model can process per second and is measured through frames per second (FPS). As a result, this metric determines whether a device can be used in real-time. The FPS for real-time object detection will vary per application. Given that drone footage involves fast motion events that can induce motion blur, 25-30 FPS provides an optimal standard for real-time object detection. For this work, power measurements are collected during inference at a sampling rate of 100 Hz using a FNIRSI FNB58 USB tester and then averaged. Lastly, a metric known as EDP is used to evaluate overall efficiency. EDP evaluates energy usage as well as application runtime since both low energy and fast runtime are beneficial for energy constrained devices [30]. This metric is calculated by taking the product of normalized energy used and normalized runtime [30]. For this work, energy and runtime are normalized to the maximum energy used and maximum runtime within a given set of comparisons. In turn, the calculation for EDP is as shown in Eqn. (5).

$$EDP = \left(\frac{Energy\ Used}{Max\ Energy\ Used}\right) * \left(\frac{Runtime}{Max\ Runtime}\right) \quad (5)$$

To evaluate runtime, all models were compared on the time required to complete inference on the test split of the AFSE dataset.

### D. Transfer Learning

Transfer Learning (TL) involves training a model on a large source dataset so that the learned weights and filters could be used as new starting points when training the same model on a much smaller target dataset. Depending on the similarity of the source and target datasets, TL can further be classified as *homogenous* or *heterogeneous*. TL as applied in this work consisted of two steps. First, a deep learning model is trained from scratch using a source dataset. Afterwards, training is continued on the target data utilizing much lower learning rates, in a process known as *Fine-Tuning*, after either freezing a specified number of layers or keeping all layers unfrozen. Fig. 2 provides an overview of the TL process. AFSE served as the target dataset whereas FASDD and COCO served as homogenous and heterogeneous source datasets, respectively.

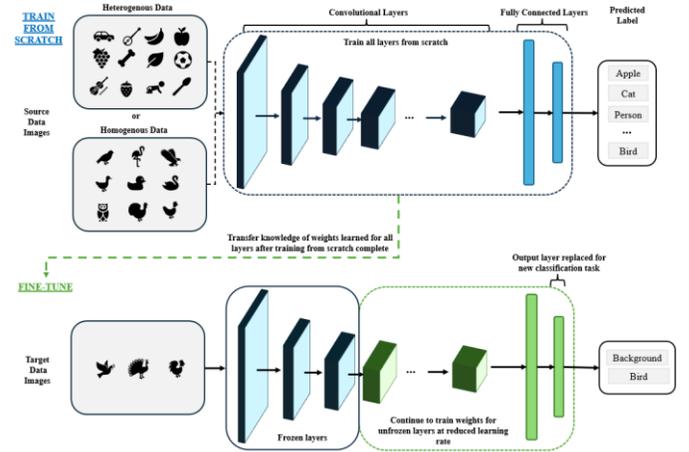

Fig. 2. Transfer learning process

### E. Model Selection

TABLE IV. YOLOv5 SIZE COMPARISON

| Model | Depth Multiple | Width Multiple | No. of Layers |
|---|---|---|---|
| YOLOv5n | 0.33 | 0.25 | 157 |
| YOLOv5s | 0.33 | 0.50 | 214 |
| YOLOv5m | 0.67 | 0.75 | 291 |
| YOLOv5l | 1.0 | 1.0 | 368 |
| YOLOv5x | 1.33 | 1.25 | 445 |

Past YOLO models have proven to be efficient and accurate for the task of flame and smoke detection [3]. Each YOLO architecture contains variants that differ in layer depth and width [31]. For instance, YOLOv5 utilizes the following sizes: nano (n), small (s), medium (m), large (l), and extra-large (x). These variants differ in the number of layers and channels as set by the parameters Depth Multiple and Width Multiple. As an example, the values for these parameters and the total number of layers for YOLOv5 are shown in TABLE IV. It has been shown with the COCO dataset that larger models tend to yield improved detection precision at the cost of speed [7]. TABLE V. lists the number of parameters (in millions) and floating points operations (FLOPs in billions) for versions 5, 6, 8, 9, 10, and 11 [7][32][33][25][26][34] respectively. Within the set of existing YOLO architectures, YOLOv5n is selected as the focus of this work due to its reduced complexity. YOLOv8n and YOLO11n are also selected as lightweight models for comparison since these are in the YOLO series developed by Ultralytics and in turn directly improve upon their prior developments. To further evaluate the effectiveness of lightweight YOLO models, a comparison against other SOTA methods for one-stage, multi-stage, and transformer-based detection models was performed. Additional models evaluated include RTM-DET, Dynamic-RCNN, and DINO. These models were implemented using an open-source object detection toolbox known as MMDetection [35]. Non-YOLO models selected were based on performance compared to other SOTA methods and availability within the model zoo of MMDetection. TABLE VI. provides a summary of these methods in terms of the number of parameters and FLOPs. A review of TABLE V. and TABLE VI. illustrate that lightweight YOLO models are less complex in terms of number of parameters and FLOPs than other SOTA methods, making them suitable for applications on edge computing devices with limited computational resources and battery capacity – such as UAVs.

TABLE V. YOLO ARCHITECTURE COMPARISON

| Size | Version | Params (M) | FLOPs (B) |
| --- | --- | --- | --- |
| n (t) | 5 \| 6 | 1.8 \| 4.7 | 5.1 \| 11.4 |
|  | 8 \| 9 | 3.2 \| 2.0 | 8.7 \| 7.7 |
|  | 10 \| 11 | 2.3 \| 2.6 | 6.7 \| 6.5 |
| s | 5 \| 6 | 7.2 \| 18.5 | 16.5 \| 45.3 |
|  | 8 \| 9 | 11.2 \| 7.2 | 28.6 \| 26.7 |
|  | 10 \| 11 | 7.2 \| 9.4 | 21.6 \| 21.5 |
| m | 5 \| 6 | 21.2 \| 34.9 | 49.0 \| 85.8 |
|  | 8 \| 9 | 25.9 \| 20.1 | 78.9 \| 76.8 |
|  | 10 \| 11 | 15.4 \| 20.1 | 69.1 \| 68.0 |
| l (c) | 5 \| 6 | 46.5 \| 59.6 | 109.1 \| 150.7 |
|  | 8 \| 9 | 43.7 \| 25.5 | 165.2 \| 102.8 |
|  | 10 \| 11 | 24.2 \| 25.3 | 120.3 \| 86.9 |
| x (e) | 5 \| - | 86.7 \| --- | 205.7 \| --- |
|  | 8 \| 9 | 68.2 \| 58.1 | 257.8 \| 192.5 |
|  | 10 \| 11 | 29.5 \| 56.9 | 160.4 \| 194.9 |

TABLE VI. SOTA OBJECT DETECTORS ARCHITECTURE COMPARISON*

| Category | Model | Backbone | Params (M) | FLOPs (B) |
| --- | --- | --- | --- | --- |
| One-Stage Detectors | RTMDET-tiny | CSPNeXt | 4.9 | 8.03 |
|  | TOOD | ResNet50 | 32.2 | 181 |
| Two-Stage Detectors | Dynamic-RCNN | ResNet50 | 41.8 | 187 |
|  | Cascade-RCNN | ResNet50 | 69.4 | 215 |
| DETR Detectors | DAB-DETR | ResNet50 | 43.7 | 92 |
|  | DINO-4Scale | ResNet50 | 47.7 | 249 |

*Params and FLOPs for this table are found using MMDetection's get_flops.py script which is listed as an experimental tool for determining these values.

## IV. EXPERIMENTAL RESULTS

To evaluate the performance of a lightweight YOLO model using the TL method on the AFSE dataset, the following experiments have been conducted:
1) Comparison of different source datasets using TL;
2) Impact of TL on generalizability of a model;
3) Impact of Cascaded TL with various datasets.
4) Comparison with other SOTA methods using TL;
5) Comparison of lightweight YOLO versions using TL.

Note that the results for experiment (5) were produced by the edge computing device while all other results were produced by the GPU server.

### A. Model Setup

TABLE VII. shows the hyperparameters used for the YOLO models of interest, all other hyperparameters not explicitly mentioned are left at default. All training was performed utilizing mini-batch learning.

Note that the number of epochs and initial learning rate varied based on the model evaluated as well as the pre-trained data used for TL. This was done to limit overfitting for each model evaluated. The AFSE dataset was split to have the following percentage ratios for the training, validation, and testing sets: 70,15,15, respectively. This split was used for all testing unless stated otherwise.

TABLE VII. YOLO NANO HYPERPARAMETER SETTINGS

| Hyperparameter | Value |
| --- | --- |
| Batch size per GPU | 16 |
| Image size | 640 |
| Epochs: YOLOv5n, COCO \| FASDD | 300 \| 150 |
| Epochs: YOLOv8n / YOLO11n, COCO \| FASDD | 150 \| 75 |
| Initial learning rate (lr0): YOLOv5n Fine-Tuning | 0.001 |
| Initial learning rate (lr0): YOLOv8n/11n Fine-Tuning | 0.0001 |

### B. Experimental Results

*1) Influence of TL Using Different Source Dataset*

TABLE VIII. shows the AP and mAP validation and testing results on the AFSE dataset after training from scratch, using TL with COCO as the source data, and using TL with FASDD as the source data. Training from scratch had been done for 150, 300, and 600 epochs. 150-epochs was included to provide a direct comparison with each step of TL. 300-epochs was included to show the time required to reach comparable accuracy. Lastly, 600-epochs was included to provide a best-case scenario. Fig. 3 provides the mAP training progression when training from scratch or utilizing TL and effectively captures how TL provides an immediate boost to accuracy. From TABLE VIII. Fine-Tuning is shown to provide improvements in AP that are more pronounced when using less frozen layers. The AP is the highest when using TL with zero frozen layers and starting from FASDD pre-trained weights. The second and sixth columns of TABLE VIII. highlight the training times required to train from scratch and perform Fine-Tuning, respectively. These results provide three key takeaways. First, it is clear that TL applied with Fine-Tuning significantly improves the detection precision compared with training from scratch on the same epoch setting. The improvement of TL with FASDD (homogeneous) is higher than that of TL with COCO (heterogeneous). By increasing to 600 epochs, training from scratch can outperform TL with COCO, but cannot reach the performance of TL with FASDD. Second, they highlight that TL helps reduce training time if no additional time is required to obtain the pre-trained weights. In fact, Fig. 3 shows that even up to 150 epochs, using TL, even with heterogenous pre-trained

TABLE VIII. YOLOv5n ACCURACY COMPARISON BETWEEN OBJECT DETECTION TRAINING SCENARIOS

| Pre-trained Weights | Training Time for Weights (Hours) | Training Description | Frozen Layers | Epochs | Training Time (Hours) | Validation | | | Testing | | |
|---|---|---|---|---|---|---|---|---|---|---|---|
| | | | | | | $AP_{fire}$ (%) | $AP_{smoke}$ (%) | mAP@0.5 (%) | $AP_{fire}$ (%) | $AP_{smoke}$ (%) | mAP@0.5 (%) |
| - | - | Train from scratch | - | 150 | 0.037 | 21.4 | 75.0 | 48.2 | 24.8 | 66.7 | 45.7 |
| | | | | 300 | 0.072 | 35.8 | 82.8 | 59.3 | 47.1 | 76.7 | 61.9 |
| | | | | 600 | 0.143 | **48.3** | **87.4** | **67.9** | **56.9** | **81.5** | **69.2** |
| COCO | - | Fine Tune | 0 | 300 | 0.071 | **40.9** | **83.0** | **61.9** | **49.7** | **80.0** | **64.8** |
| | | | 5 | | 0.067 | 31.2 | 80.2 | 55.7 | 39.2 | 78.1 | 58.6 |
| | | | 10 | | 0.064 | 27.0 | 69.7 | 48.4 | 30.6 | 67.5 | 49.1 |
| FASDD | 9.604 | Fine Tune | 0 | 150 | 0.037 | **56.2** | **92.0** | **74.1** | **70.0** | **88.5** | **79.2** |
| | | | 5 | | 0.034 | 53.0 | 91.1 | 72.1 | 63.4 | 88.3 | 71.8 |
| | | | 10 | | 0.031 | 53.8 | 85.8 | 69.8 | 53.9 | 78.6 | 64.2 |

TABLE IX. YOLO NANO TL ACCURACY COMPARISONS

| Pre-trained Weights | Model | Validation | | | Testing | | |
|---|---|---|---|---|---|---|---|
| | | $AP_{fire}$ (%) | $AP_{smoke}$ (%) | mAP@0.5 (%) | $AP_{fire}$ (%) | $AP_{smoke}$ (%) | mAP@0.5 (%) |
| Train From Scratch | v5n | 35.8 | **82.8** | **59.3** | **47.1** | 76.7 | 61.9 |
| | v8n | 36.4 | 81.3 | 58.8 | 46.5 | **85.2** | **65.8** |
| | 11n | **36.7** | 79.5 | 58.1 | 37.7 | 83.2 | 60.4 |
| COCO | v5n | 40.9 | 83.0 | 61.9 | 49.7 | 80.0 | 64.8 |
| | v8n | 45.9 | **91.5** | 68.7 | 54.0 | **87.7** | 70.9 |
| | 11n | **50.8** | 88.1 | **69.4** | **60.6** | 86.2 | **73.4** |
| FASDD | v5n | 56.2 | 92.0 | 74.1 | **70.0** | 88.5 | **79.2** |
| | v8n | **57.6** | 93.4 | **75.5** | 59.6 | **94.0** | 76.8 |
| | 11n | 52.7 | **95.9** | 74.3 | 63.9 | 91.1 | 77.5 |

weights, is able to outperform training from scratch. Lastly, the results of TABLE VIII. capture that slight improvements in speed when using TL could be obtained by increasing the number of frozen layers with the tradeoff of reduced precision.

A similar comparison is again performed, this time for all Ultralytics' YOLO nano variants, the results of which are shown in TABLE IX. Note that for these experiments, no layers were frozen during Fine-Tuning since it had been shown in TABLE VIII. that Fine-Tuning without freezing any layers yields the best performance. The YOLOv5n results are repeated for reader clarity. These comparisons illustrate the same trend shown previously in which TL can help improve model AP and that Fine-Tuning with a homogenous source dataset yields the largest performance improvement.

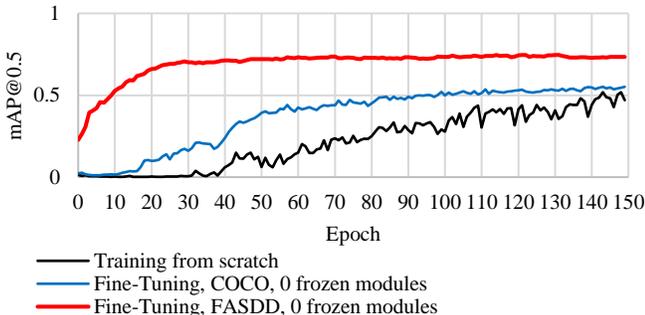

Fig. 3. Training accuracy comparison

*2) Influence of TL on Generalizability*

Generalizability was evaluated using Stratified $k$-Fold Cross-Validation and the standard deviation of AP. Note that stratified $k$-fold is used over standard $k$-fold since there is an imbalance between fire and smoke instances and in turn this same proportion is sought after in each split [36]. Cross-validation is utilized to take into account the ways different splits on the same dataset impact the training and validation results of a model. When implementing standard $k$-fold cross validation, a dataset is separated into $k$-splits, or folds. Given the $k$-folds, one split is used for validation while the rest are used for training. Training and validation are repeated $k$-1 times, with each iteration using a different split from the folds produced for validation. For this experiment, 5-folds are used rather than 10 due to the small size of the AFSE dataset. From Fig. 4, it is evident for both the fire and smoke classes that variance is reduced after TL is applied when compared to training from scratch for 150 epochs. However, the reduction in variance was more notable when training for 600 epochs. This result confirms that application of TL can make a model less susceptible to fluctuations in real-world data. Moreover, this experiment augments the observation that the use of TL when using homogeneous pre-trained weights can reach higher APs than training from scratch.

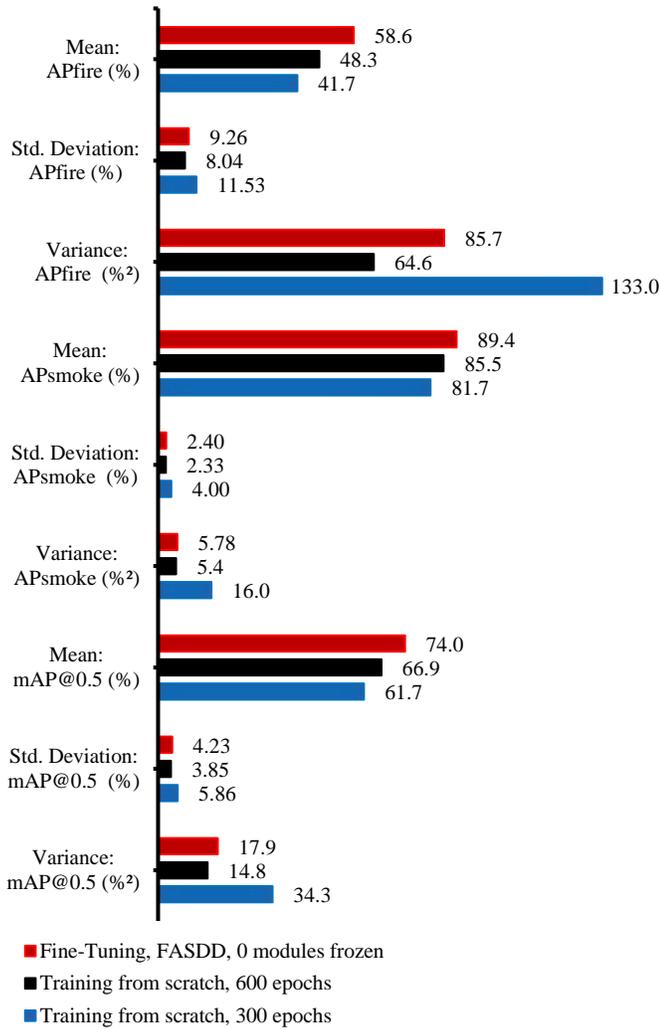

Fig. 4. Five-Fold Cross-Validation generalization comparison

*3) Influence of Cascaded TL*

A comparison implementing a cascaded approach to TL is applied to determine the effectiveness of training a model on multiple datasets. A summary of the results is shown in TABLE X. Three approaches are implemented towards this end. In the first approach, a YOLOv5n model is trained on FASDD starting from COCO pre-trained weights. This is done with 0 layers frozen and 10 layers frozen for 150 epochs in both instances. The YOLOv5n model is then trained for an additional 150 epochs on the AFSE dataset, again for 0 and 10 frozen layers. The second approach followed this same methodology with the difference being that the first stage in the cascade of the model is trained on the D-Fire dataset starting from FASDD pre-trained weights. Note that in this instance, the FASDD pre-trained weights have been trained from scratch for 150 epochs. The third approach serves to compare against a model trained without cascading by merging the D-Fire and FASDD datasets. This aggregated dataset was used to train a model from scratch for 150 epochs. From these experiments, four comparisons can be performed.

The first comparison is between models with a different number of frozen layers in stage one using the same pre-trained weights and number of frozen layers in stage two. For the models starting from COCO pre-trained weights, AP for each class tends to drop when the backbone is frozen in stage one. When starting from FASDD pre-trained weights, AP for each class tends to increase when the backbone is frozen in stages one and two but only fire AP increases when no layers are frozen in either stage. For the models trained utilizing COCO pre-trained weights, the drop in performance after freezing the backbone in stage one is the result of not being able to alter irrelevant early features learned. As for the models utilizing FASDD pre-trained weights, the performance improvement is the result of freezing useful early features learned and continued fine-tuning on relevant datasets. When unfreezing all layers in stage one for the FASDD case, useful features learned were lost in attempting to learn the D-FIRE dataset.

The second comparison is between models starting from different pre-trained weights but using the same number of frozen layers in stages one and two. From this comparison, AP tends to be worse when using zero frozen layers in stage one and starting from FASDD. However, this same comparison yields mixed results when zero layers are frozen in stage two. In this instance, AP is worse for the validation data but yields improved AP on the test data. A different pattern is noticed when the backbone is frozen in stage one. In this case, starting from homogeneous pre-trained weights and fine-tuning with additional homogenous datasets tends to yield better results than starting from heterogeneous pre-trained weights and fine-tuning with homogeneous datasets. These results support the same notion inferred from the first comparison. Moreover, they indicate that when starting from a large relevant dataset, performance can be hampered when fine-tuning in a cascaded fashion on a smaller, even if still relevant, dataset.

The third comparison is between a model trained using a merged D-Fire and FASDD dataset to a model trained using a cascaded approach, with the same datasets, for varying numbers of frozen layers in stage one. When comparing the merged model to the cascaded model starting from FASDD for the zero frozen layer stage one training case, AP generally improves when starting from the merged dataset pre-trained weights. A similar pattern is noticed in the test data when starting from FASDD for the backbone frozen stage one training case. The validation results for this case are mixed. The final item compared between the merged and cascaded approach is the training time. Here, after adding all training time required when using the cascaded approach, training with the merged datasets takes less time.

Lastly, when comparing the results between TABLE VIII. and TABLE X. it is noticed that an additional stage of TL, even when using relevant datasets for all training stages, yields worse or similar results to the models only applying one stage of TL. In turn, cascaded TL, as applied in this work, does not yield any notable benefits. Moreover, it is shown that the best training results for a model using TL are obtained by using the merged FASDD and D-Fire data to develop the pre-trained weights.

However, the inclusion of D-Fire was only able to add marginal performance gains in the best-case scenario with the overhead of extra training time.

*4) Comparison with SOTA Models*

Comparisons performed for YOLOv5n against SOTA detectors are summarized in TABLE XI. To maintain a fair comparison, all models are trained on the AFSE dataset starting with pre-trained FASDD weights. All hyperparameters, excluding batch size and number of epochs, are left at the defaults configured within MMDetection [35]. The number of epochs utilized for models differs to ensure each model reaches a stable mAP value. Batch size is also modified for several models due to memory limitations when training. The results for YOLOv5n, YOLOv8n, and YOLO11n are repeated for reader clarity. Dynamic-RCNN and YOLOv8n yield the best performance for the validation data, both achieving a 75.5 mAP. However, YOLOv5n yields the best performance for the test data, achieving a 79.3 mAP. Consequently, it is evident that YOLOv5n remains competitive as an object detection model.

A visual comparison is provided in Fig. 5. These images are selected to provide a variety of fire and smoke instance examples. Column one shows a small fire instance and a translucent smoke instance. Column two shows large prominent fire and smoke instances. Column three contains no smoke nor fire but does contain small regions with colors like those produced by fire and smoke. Column four contains a medium sized fire instance and a range of transparencies/sizes for the smoke instances. Lastly, column five contains no smoke nor fire but has regions with clouds to mimic smoke. The following observations are noted for these comparisons. Column one shows that all models except RTM-DET Tiny, YOLOv8n, and YOLO11n can capture the true positives. These three models produce false negatives for the fire instance. Column two demonstrates all models can capture the true positives, although with varying confidence scores. Sometimes, such as for Dynamic-RCNN in column one, the smoke instance is incorrectly labeled as multiple true positives. For column three, YOLOv8n and RTM-DET Tiny produce false positives. For column four, all models except RTM-DET Tiny and DINO produce a false negative by missing the leftmost smoke instance. Lastly, for column five, all models except YOLOv5n, YOLO11n, and DINO produce false positives. Overall, YOLOv5n performs the best in terms of fire detection and comparable in detecting smokes

TABLE X. YOLOv5n CASCADED TL COMPARISON

| Pre-trained Weights | Weights Train Time (Hours) | Stage 1 Train Dataset | Layers Frozen Stage 1 | Stage 1 Train Time (Hours) | Stage 2 Train Dataset | Layers Frozen Stage 2 | Stage 2 Train Time (Hours) | Validation | | | Testing | | |
|---|---|---|---|---|---|---|---|---|---|---|---|---|---|
| | | | | | | | | $AP_{fire}$ (%) | $AP_{smoke}$ (%) | mAP@0.5 (%) | $AP_{fire}$ (%) | $AP_{smoke}$ (%) | mAP@0.5 (%) |
| COCO | - | FASDD | 0 | 9.268 | AFSE | 0 | 0.037 | **57.7** | **93.1** | **75.4** | **68.1** | **87.9** | **78.0** |
| | | | | | | 10 | 0.031 | 56.1 | 85.3 | 70.7 | 56.0 | 81.0 | 68.5 |
| COCO | - | FASDD | 10 | 9.589 | AFSE | 0 | 0.036 | **46.9** | **91.8** | **69.3** | **58.9** | **85.6** | **72.3** |
| | | | | | | 10 | 0.032 | 42.1 | 83.7 | 62.9 | 50.7 | 77.8 | 64.3 |
| FASDD | 9.604 | DFIRE | 0 | 1.741 | AFSE | 0 | 0.035 | **52.0** | **91.8** | **71.9** | **69.5** | **91.4** | **80.5** |
| | | | | | | 10 | 0.032 | 50.9 | 84.7 | 67.8 | 55.0 | 82.7 | 68.9 |
| FASDD | 9.604 | DFIRE | 10 | 1.618 | AFSE | 0 | 0.036 | **55.3** | **90.1** | **72.7** | **69.9** | **89.8** | **79.9** |
| | | | | | | 10 | 0.032 | 55.1 | 86.0 | 70.6 | 57.8 | 82.3 | 70.1 |
| FASDD DFIRE MERGED | 11.004 | AFSE | 0 | 0.036 | - | - | - | **57.9** | **94.1** | **76.0** | **70.1** | **90.6** | **80.3** |
| | | | 10 | 0.031 | | | | 52.7 | 85.1 | 68.9 | 58.3 | 82.7 | 70.5 |

TABLE XI. SOTA DETECTION MODELS COMPARISON

| Category | Model | Batch Size Per GPU | Epochs | Validation | | | Testing | | |
|---|---|---|---|---|---|---|---|---|---|
| | | | | $AP_{fire}$ (%) | $AP_{smoke}$ (%) | mAP@0.5 (%) | $AP_{fire}$ (%) | $AP_{smoke}$ (%) | mAP@0.5 (%) |
| One-Stage | RTM-DET Tiny | 4 | 303 | 57.0 | 89.4 | 73.2 | 69.2 | 87.6 | 78.4 |
| | YOLOv5n | 16 | 150 | 56.2 | 92.0 | 74.1 | **70.0** | 88.5 | **79.3** |
| | YOLOv8n | 16 | 75 | 57.6 | 93.4 | **75.5** | 59.6 | **94.0** | 76.8 |
| | YOLO11n | 16 | 75 | 52.7 | **95.9** | 74.3 | 63.9 | 91.1 | 77.5 |
| Two-Stage | Dynamic-RCNN | 4 | 122 | **64.1** | 86.8 | **75.5** | 64.8 | 84.9 | 74.9 |
| Transformer | DINO-4scale | 1 | 23 | 57.6 | 87.1 | 72.4 | 69.5 | 88.7 | 79.1 |

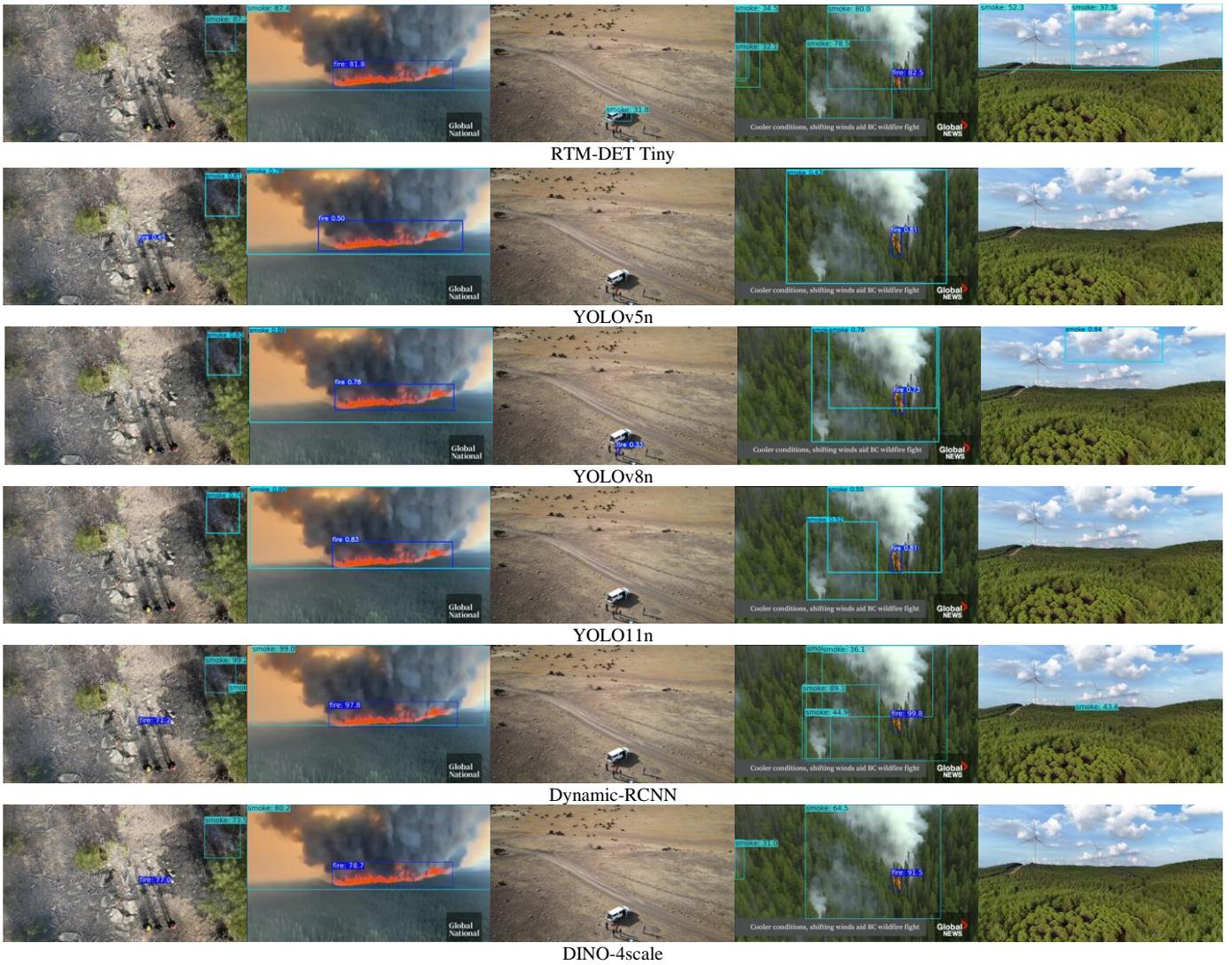

Fig. 5. Comparison of inference results on five scenarios after training on AFSE data and having started from pre-trained FASDD weights. From left to right, columns 1-5.

*5) Comparison of Lightweight YOLO Versions on Edge Devices*

Using the GPU configuration, detailed in TABLE III. all lightweight YOLO models are trained from scratch for 300 epochs using default hyperparameters with a batch size of 16 per GPU (96 total). The models are then evaluated on the edge computing device using the configuration detailed in TABLE II. The validation accuracies and FPS for each model of interest are shown in TABLE XII. The EDP for these models is provided in Fig. 6. Note that each model's EDP was normalized to the largest the largest runtime and energy consumed within the comparisons of TABLE XII. From the results, YOLOv5s achieves the best validation accuracy while YOLOv5n is shown to yield the highest FPS and the lowest EDP followed by YOLOv8n and YOLO11n. When evaluated against the other comparably complex models, these three lightweight architectures yield higher mAPs. As a result, these models are selected for further evaluation using TL.

TABLE XIII. lists the testing accuracies, FPS, and average power for YOLO nano versions 5, 8, and 11 after training with TL having started from COCO or FASDD pre-trained weights. Training from scratch results were also repeated in this table for ease of comparison. The EDP for these models is provided in Fig. 7. Each model's EDP was normalized to the largest runtime and energy consumed within the comparisons of TABLE XIII. From these results, it is evident that each model evaluated benefits from TL with improved AP. However, regarding which model yields the best AP, mixed results are observed whether starting from scratch, COCO, or FASDD pre-trained weights. Notably, the FPS for YOLOv5n is found to be nearly double that of either YOLOv8n or YOLO11n. Moreover, although YOLOv5n on average uses more power during inference, the overall amount is negligible. Additionally, the EDP for YOLOv5n is found to yield the best results regardless of whether training from scratch or starting from pre-trained weights. Notably, none of the YOLO models seem to have improved FPS, power usage, or normalized EDP after utilizing TL when compared to the same model trained from scratch.

TABLE XII. PERFORMANCE COMPARISON, LIGHTWEIGHT YOLO MODELS

| YOLO Model Trained from Scratch | Avg. FPS | Avg. Power During Inference (mW) | Validation | | |
|---|---|---|---|---|---|
| | | | $AP_{fire}$ (%) | $AP_{smoke}$ (%) | mAP@0.5 (%) |
| YOLOv5s | 2.8 | 7713.53 | 44.0 | 86.1 | **65.0** |
| YOLOv6s | 1.0 | **6534.68** | 37.3 | 85.4 | 61.3 |
| YOLOv8s | 1.2 | 6702.58 | **45.5** | 83.5 | 64.5 |
| YOLOv9s | 1.1 | 6760.80 | 40.3 | **89.2** | 64.7 |
| YOLOv10s | 1.1 | 6758.71 | 37.0 | 77.3 | 57.2 |
| YOLO11s | 1.3 | 6764.85 | 42.6 | 86.3 | 64.4 |
| YOLOv5n | **5.9** | 6783.22 | 35.8 | **82.8** | **59.3** |
| YOLOv6n | 3.1 | **6374.28** | 35.1 | 82.3 | 58.7 |
| YOLOv8n | 3.3 | 6522.63 | 36.4 | 81.3 | 58.8 |
| YOLOv9t | 2.7 | 6593.53 | 36.3 | 79.5 | 57.9 |
| YOLOv10n | 2.7 | 6630.84 | 21.1 | 72.3 | 46.7 |
| YOLO11n | 3.3 | 6580.66 | **36.7** | 79.5 | 58.1 |

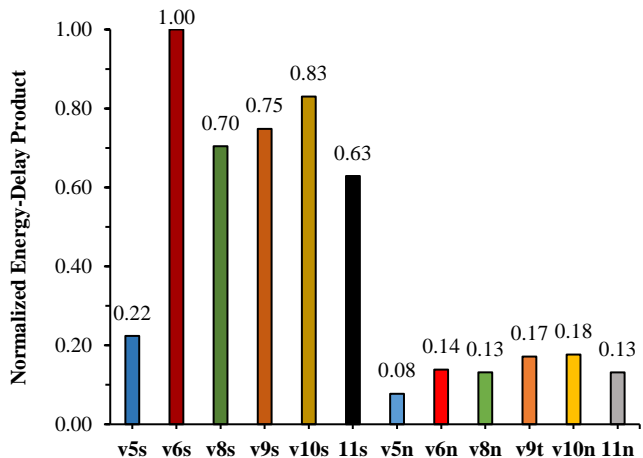

Fig. 6. Normalized energy-delay product comparison for lightweight YOLO models when training from scratch.

TABLE XIII. PERFORMANCE COMPARISON, LIGHTWEIGHT YOLO MODELS WITH TL

| Pre-trained Weights | YOLO Model | Avg. FPS | Avg. Power During Inference (mW) | Testing | | |
|---|---|---|---|---|---|---|
| | | | | $AP_{fire}$ (%) | $AP_{smoke}$ (%) | mAP@0.5 (%) |
| Train From Scratch | v5n | **6.1** | 6924.63 | **47.1** | 76.7 | 61.9 |
| | v8n | 3.3 | **6525.87** | 46.5 | **85.2** | 65.8 |
| | 11n | 3.3 | 6560.33 | 37.7 | 83.2 | 60.4 |
| COCO | v5n | **6.1** | 6895.64 | 49.7 | 80.0 | 64.8 |
| | v8n | 3.3 | **6515.34** | 54.0 | **87.7** | 70.9 |
| | 11n | 3.3 | 6561.99 | **60.6** | 86.2 | **73.4** |
| FASDD | v5n | **6.1** | 6886.54 | **70.0** | 88.5 | **79.2** |
| | v8n | 3.3 | **6542.87** | 59.6 | **94.0** | 76.8 |
| | 11n | 3.2 | 6569.48 | 63.9 | 91.1 | 77.5 |

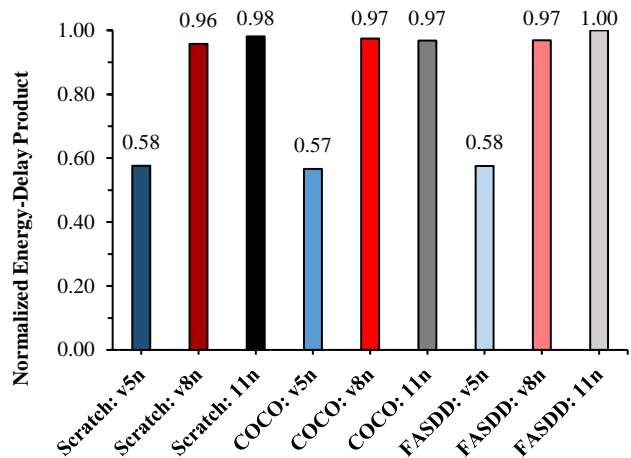

Fig. 7. Normalized energy-delay product comparison for YOLO models v5n, v8n, and v11n when fine-tuning from either COCO or FASDD source dataset.

## V. CONCLUSIONS

Unmanned Aerial Vehicles (UAVs) equipped with deep learning (DL)-enabled computer vision are emerging as an indispensable tool in the early detection of wildfires, particularly when enhanced by Transfer Learning (TL) to address challenges of data scarcity. The application of homogeneous TL significantly improves the accuracy, efficiency, and generalizability of aerial-based wildfire detection models. Although extending training epochs may not directly increase accuracy, it does effectively broaden the model's applicability across diverse scenarios.

The study reveals that cascaded TL, despite involving multiple datasets, does not consistently surpass the performance of a simpler, single-stage TL approach. This finding underscores that greater complexity in TL frameworks does not necessarily yield superior results. Moreover, the most effective strategy for leveraging multiple large datasets is to amalgamate them into a robust pre-training base, thereby optimizing the learning trajectory.

Regarding operational metrics, TL demonstrates minimal impact on inference times, power usage, and energy consumption when deployed on edge computing devices. This ensures that operational efficiency is maintained without compromise. Notably, the YOLOv5 architecture continues to prove its worth for implementation on edge devices, effectively competing with other state-of-the-art models and highlighting its adaptability to real-time, resource-constrained environments.

Looking forward, future research will focus on further refining TL-enhanced models to maximize their effectiveness for real-time applications on edge computing devices. This strategic emphasis aims to improve operational efficiencies and expand the capabilities of wildfire detection systems, ultimately enabling faster and more effective responses to these critical natural events.

ACKNOWLEDGMENTS

This work is supported in part by the Nevada Space Grant Consortium Research Opportunity Fellowship, NSF under grant no. 1949585, and the UNLV AI SUSTEIN Seed Grant.